# A Calibrated Instrument for Measuring How Inference Optimizations Affect Output Quality

Jerry Kaplan[1]
Adjunct Lecturer, Computer Science and Masters in International Policy
Stanford University

*Draft V1 of September 11, 2026*

## Abstract

Large language model optimization is an area of intense research, with distinct branches ranging from quantization of model weights, to early-exit methods for skipping model layers, to speculative decoding – which uses small models to propose tokens and larger models for verification. The published literature for each of these tracks uses its own quality measures — typically an idiosyncratic benchmark score. Few approach the degree of measurement precision required by other scientific disciplines.

We propose a rigorous methodology for measuring the quality of model output, suitable for cross-system and cross-technique comparison. We score outputs by using an LLM as a judge (a common practice), but we calibrate and apply the judge using formal methods: To calibrate, we compare the judge's scores on two ordinary runs of a model given the same sequence of prompts, to verify that the judge shows no systematic preference between statistically equivalent outputs, and to measure its per-sample noise. Then as part of any experimental design, we include one 'null' condition — provably identical in distribution to the unmodified model — whose measured difference must therefore be zero: a nonzero reading flags a defect in the experimental design.

With this one calibrated instrument we measure a collection of acceleration techniques on the same prompts, so their output quality costs can be meaningfully compared. We also show that perceived quality is highly dependent on the domain of discourse. For example, we found that running a 4-bit precision model instead of its 16-bit (full precision) variant produced output statistically indistinguishable from the original down to the ±0.3-point resolution of our design, on English prose and on Chinese alike. One step further, at 3-bit precision, the same test prompts lost 0.5 points in English prose but 0.9 points in Chinese and 1.1 points on multi-step math problems; and an early-exit configuration that lost 0.7 points on English prose lost 2.5 points on multi-step math problems, cutting the number of correctly solved problems from 19 of 27 to 6. This pattern held whether the model was developed by a Chinese company (Alibaba) or a U.S.

[1]Corresponding author: Jerry Kaplan, Adjunct Lecturer, Computer Science and Masters in International Policy, Stanford University.

company (Meta), but its magnitude did not: the same 3-bit quantizer cost Meta's model 1.8 points where it cost Alibaba's 0.7.

We also found that a model's certainty about a token predicts how likely that token is to differ from the full model's choice, but is nearly unrelated to how much that difference would impact the judged quality of the output. Changing a high-certainty comma to a semicolon, for instance, may have little effect, while changing one high-certainty digit for another can render an entire response factually incorrect. Acceptance rules that rely on certainty alone therefore cannot distinguish errors that matter from errors that don't.

These results demonstrate the need for a standardized and rigorous regime for evaluating and comparing lossy optimization techniques across potential deployment scenarios.

## 1. Introduction

Most optimizations that make a language model cheaper to run change its output. Weight quantization changes every logit. Early exit computes some positions with fewer layers than the model was trained to use. Speculative decoding is the exception when it is run with the standard rejection rule: a small draft model proposes several tokens, the full model verifies them in one pass, and the acceptance rule guarantees that every emitted token is distributed exactly as the full model would have produced it (Leviathan et al., 2023; Chen et al., 2023). That guarantee holds only for the strict rule; the relaxed acceptance rules proposed to raise throughput further (Bachmann et al., 2025; Xia et al., 2026) trade it away, and become lossy optimizations like the others. The question for the practitioner, in every lossy case, is whether the output is worse, and by how much, on the material that deployment will see. The question is easy to state and, as commonly answered, easy to get wrong.

The standard answers are proxies. Perplexity on held-out text measures how well the compressed model predicts the original model's tokens, not whether its own generations are any good; we observe below that a degenerate, repetitive output can score lower perplexity than the reference it replaced. Token-level agreement — the fraction of positions where the compressed and full models choose the same next token — is the natural quantity for speculative decoding, since it is the expected acceptance rate, but it averages over positions of very different consequence: a wrong whitespace token and a wrong digit count the same. Benchmark accuracy is closer to what matters, but Dutta et al. (2024) showed that equal accuracy can hide large numbers of answers that flip from right to wrong and back, and Kübler et al. (2026) showed that per-sample paired tests detect degradations that aggregate accuracy misses. None of these measures says whether an eight-hundred-token explanation, program, or proof got worse.

A large language model used as a judge can answer that question, and is increasingly used for this purpose. But a judge is a measuring instrument, and the optimization literature has so far used it without the calibration any other instrument would require: a demonstration that it reads zero when the true difference is zero, a demonstration that it reads a known loss with the right sign and a plausible size, a stated resolution, and a stated smallest effect of interest. Usami et al. (2026) recently made the general case for characterizing judges this way, using synthetic null inputs; Helcig et al. (2026) defined "statistically lossless" quantization relative to the uncompressed model's own run-to-run variance on benchmark tasks. This paper combines those ideas with two others and applies the result across three optimization families on identical prompt sets.

The design has five parts. First, a paired design with a dual reference: the unmodified model is sampled twice for each prompt, and every optimized condition — we call each condition an arm — is compared with the mean of the two. Second, an exchangeability null: the two unmodified samples are exchangeable, so the judged difference between them has expectation exactly zero, and its spread is the instrument's noise, measured on the same prompts with the same judge as the arms. Third, an implementation null: a speculative-decoding arm under the strict rejection rule, whose emitted distribution equals the full model's by theorem, so that any nonzero reading indicts the experimental implementation rather than the technique being tested. Fourth, positive controls: arms whose losses were measured in earlier runs, so that the instrument's sensitivity at its operating point is demonstrated rather than assumed. Fifth, a pre-specified equivalence bound of 0.3 rating points (on a 7-point scale; chosen before the run as the smallest difference of practical interest), tested with two one-sided tests. Two of our chosen test domains permit comparison to factual ground truth (i.e. is a computed answer correct?), so we report those results in addition to the judge's ratings.

Applied to the Qwen2.5-7B-Instruct publicly available model, the selected test arms were quantization (NF4 4-bit, HQQ 3-bit – two arms), speculative decoding (strict and lenient acceptance with a 4-bit drafter – two arms), and early exit (layer 15 with a fitted cache-repair map, with and without periodic exact refill – two arms). On 220 prompts across five different subject-matter domains, the instrument gives a quality map with confidence intervals of about ±0.1 rating points pooled and ±0.3 per domain. Four of the test arms proved equivalent to the full model within the ±0.3 bound — both nulls (as expected), lenient speculation, and NF4 quantization — and three are not. The lossy arms – where the judge can resolve differences in quality – show highly divergent results in different domains, illustrating why knowing the expected prompt traffic is critical when selecting optimization techniques. For example, 3-bit quantization costs 0.1 judging points on an arithmetic word problem domain but 1.1 on multi-step math problems, 10x worse. Early-exit with refill costs 0.7 judging points on a plain English prose domain but 2.5 on multi-step math problems (about 3.5x worse).

This paper's novel contribution to the field is a formal test protocol for consistently measuring and comparing differences in output quality for optimization techniques. Future work can extend this protocol to other techniques that potentially impact output quality, such as context compaction, model alignment drift, KV-cache eviction, prompt compression, retrieval truncation, safety filtering, watermarking, distillation, and changes to system-prompts. The code, probe prompts, and other materials needed to reproduce or apply this protocol are available at https://github.com/jerrykaplan/Calibrated-Instrument.

## 2. Related Work

**Evaluating compressed models.** Jaiswal et al. (2024) and Dutta et al. (2024) documented that perplexity and accuracy under-report the effects of compression; the latter introduced answer flips — benchmark items whose answer changes from correct to incorrect, or the reverse, between the baseline and the compressed model — and KL divergence as complements and used a GPT-4 judge to show that high-flip models are worse in free-form generation. Kurtić et al. (2025) evaluated quantized Llama models with Arena-Hard-Auto and text-similarity metrics, reporting means without uncertainty; Kübler et al. (2026) criticized that practice and proposed exact paired tests on per-sample correctness, noting that even provably lossless kernel changes perturb outputs through floating-point differences between kernels. Helcig et al. (2026) defined statistically lossless quantization as staying within the uncompressed model's run-to-run variance and introduced the expected acceptance rate (EAR, one minus the total-variation distance between draft and target next-token distributions) as a distribution-level fidelity measure; we adopt their name for that quantity and report it beside judged quality. Gao et al. (2024) detect quantization and watermarking as distribution shifts with two-sample tests, which is complementary: it asks whether the output distribution changed, not whether the outputs got worse.

**Language and domain dependence.** Marchisio et al. (2024) found that quantization harms non-Latin-script languages most and that automatic metrics understate the harm relative to human and LLM-judge evaluation; Borgersen and Goodwin (2025), using llama.cpp k-quantization on Llama-3.3-70B and automatic benchmarks only, found no significant multilingual effect. The two are reconcilable: a 70-billion-parameter model tolerates 4 bits far better than a 7-billion one, and automatic metrics are what Marchisio et al. found insensitive. Our per-domain intervals are consistent with both: at 4 bits we find no Chinese-specific loss on a 7-billion-parameter model; at 3 bits we find the ordering Marchisio et al. reported.

**Judges as instruments.** Miller (2024) set out the variance decomposition and interval reporting that evaluations should carry. Usami et al. (2026) proposed a datasheet for LLM judges — the response to empty or identical inputs, sensitivity on a set of deliberately degraded responses whose true ordering is known, positional bias — framed explicitly as instrument characterization. Yagubyan (2026) re-ran pairwise judgments fifty times per item and found the preferred response flips on 13.6% of trials on average. We differ in how we test the null condition: our second reference sample has real, deployment-representative variation, and its zero mean results from exchangeability rather than trivially from identical test inputs. We also differ in the positive controls, which are real optimizations with previously measured losses rather than constructed defects.

**Lossy speculative decoding.** Judge Decoding (Bachmann et al., 2025) and AutoJudge (Garipov et al., 2025) learn relaxed acceptance rules; Xia et al. (2026) survey training-free relaxations under a single relaxation parameter and observe that, unlike lossless speculation, relaxed variants require some form of quality evaluation. Our leniency arm is one member of that family, included as a measurement target. Ziashahabi et al. (2025) make the same distinction we measure, but as a mechanism rather than a measurement: they argue that matching the target model's sampling distribution is too strict a requirement, and that a decoding strategy should instead preserve the target's utility, rejecting only the "pivot" tokens whose substitution would change the final outcome. Their classifier is trained to find those tokens; our instrument asks, after the fact, how much a given acceptance rule cost. Speculative decoding with a quantized drafter of the same model has been proposed several times (QSpec, QuantSpec, ML-SpecQD: Zhao et al., 2025; Tiwari et al., 2025; Georganas et al., 2025); we use it because it's a richer null arm than simply running the unmodified model multiple times: it exercises the speculative-decoding code, so a nonzero reading indicts that code.

**Early exit.** Layer-skipping and early-exit inference (Schuster et al., 2022; Elhoushi et al., 2024; Kaplan, 2026) are usually evaluated by perplexity or by token agreement with the full model. Our earlier work (Kaplan, 2026) found that a token-level fidelity threshold can accept configurations that destroy task correctness; this paper measures the same effect with a judged instrument on whole responses.

## 3. The Instrument

### 3.1 Judge and rubric

Each response is rated in isolation by claude-sonnet-5 on a seven-point scale (7: complete, correct, and well formed; 4: a clear factual or logical error, repetition, or visible incompleteness; 1: degenerate or unusable), with two Boolean flags ("derailed" – meaning ran off topic, and "recovered" – meaning got back on track after an initial derailment), plus a written rationale of at most fifteen words. The judge sees only the prompt and the response, no other information about the test arm, etc. The rubric, as presented to the judge, is included in Appendix A.

### 3.2 Paired design with a dual reference

For each prompt the unmodified model is sampled twice at the study temperature, giving two reference samples and hence a pair of reference ratings. Every optimized arm is sampled once from the same prompt, and its score is its judged rating minus the mean of the two unmodified ratings. Scores are therefore differences on the seven-point scale. Every Δ reported below is a mean of such differences, which is why the tables are centered on zero. Averaging two references halves the contribution of the reference's own sampling noise. In a pilot with a single reference, half the variance of an arm's score was shared across arms on the same prompt (intraclass correlation 0.49), because a weak (single) reference draw raises or lowers every arm's score at once; with the dual reference that correlation is reduced to 0.12.

### 3.3 The exchangeability null

The two reference samples are independent draws from the same distribution. Exchanging them changes nothing, so the expected judged difference between them is exactly zero, for any judge, however biased that judge might be (higher or lower than a theoretically optimal judgment). The observed difference is therefore a direct measurement of the instrument's noise on the study's own prompts, and its spread (standard deviation 0.94 per prompt; 0.84 for an arm's score against the dual baseline) is the quantity from which the required number of prompts is computed. In this study the difference – sample 2 minus sample 1 – was −0.05 with a 95% confidence interval of [−0.18, +0.08] over 220 prompts. There are thus three null generations per prompt — the two reference samples and the lossless speculative arm of Section 3.4 — and two null tests: sample 2 against sample 1 (this section), and the lossless speculative arm against the reference pair (Section 3.4).

### 3.4 The implementation null

Speculative decoding under the standard rejection rule — accept a drafted token x with probability min(1, p_target(x)/p_draft(x)), and on rejection sample the replacement from the normalized residual max(0, p_target − p_draft) — emits every token from the unmodified (target) model's distribution exactly, whatever the drafter proposes. An arm generated this way with a 4-bit drafter of the same model is therefore null by construction, but its outputs pass through the draft cache, the target cache, batched verification, cache cropping, carried-over tokens, and end-of-sequence handling. If any of that is wrong, the emitted distribution is no longer the target's, and the judge reads a difference. The two null tests localize faults: if sample 2 minus sample 1 is nonzero, the judge or the analysis is biased; if that difference is zero but the lossless speculative arm minus the reference pair mean is nonzero, the fault is in the implementation. In our own development this control detected two defects that would otherwise have appeared as quality effects. On Qwen it reads +0.05 [−0.06, +0.16]; on Llama, −0.14 [−0.27, −0.01], a reading discussed in Section 5.3.

Two qualifications. First, the guarantee assumes exact arithmetic. The verifier computes its probabilities in one batched pass over the drafted window, and batched and sequential bf16 passes sum in different orders, so the verifier's probabilities differ in the last bits from those the unmodified model produces token by token. The acceptance rule — accept the drafted token with probability min(1, p_verifier/p_draft), otherwise discard it and resample from the verifier — therefore rejects a few tokens even when the drafter is the unmodified model itself: 0.7% of drafted tokens in our control runs. The null holds up to floating point resolution, not perfect resolution.

Second, the same floating-point effect makes token identity a poor standard for "unchanged" even before any optimization is applied. Generation for this study ran on an RTX 5090; judging and analysis ran on a laptop; and an earlier check ran the unmodified model on both under greedy decoding (temperature 0, so that the output is determined by the logits alone). The two machines' outputs diverged after a few hundred tokens, a difference attributable exclusively to the floating point summation order. Sampling at the study temperature adds a much larger source of variation on top of this. That variation is what the exchangeability null measures, and it is why the instrument compares distributions of judged quality rather than token sequences.

### 3.5 Positive controls

A calibration built only on nulls shows that the meter reads zero when zero is true. It should also read a known loss with the right sign and a plausible size. Two early-exit arms measured in an earlier campaign under greedy decoding serve this purpose: exit at layer 15 with cache repair and an exact recomputation of the cache (a "refill") every 16 tokens, which had measured −0.59, and the same without refill, −1.05. Both values were registered as predictions before the run. Both arms were detected with the right sign and ordering, and both were under-predicted by about a factor of two as detailed below, because the present run used sampling rather than greedy decoding and added a domain of verifiable multi-step problems. The instrument's sensitivity is demonstrated; the anchors' miss is itself information about how damage depends on the decoding regime and the material.

### 3.6 Equivalence, power, and pre-registration

"Not significantly different from zero" is not "equivalent." We state a smallest effect of interest — 0.3 rating points, the resolution of the pilot ladder and below any effect the judge flagged as noticeable in the pilots — and declare an arm equivalent to the full model when its 90% confidence interval lies inside ±0.3, which is the two-one-sided-tests procedure at $\alpha = 0.05$ (Lakens, 2017). With a score standard deviation of 0.84, a pooled interval of ±0.11 needs 220 prompts; a per-domain interval of ±0.3 needs 40 (60 for code, whose per-item standard deviation is 1.4). Resolution is a choice, not a property of the protocol: the interval narrows as the square root of the number of prompts, so a study needing ±0.05 pooled rather than ±0.11 would run about 1,100 prompts instead of 220, and one content with ±0.2 would run 70. We report the resolution we bought so that a reader can price a different one. Predictions for every arm, with confidence bands, were written into the analysis script before generation began, and the registered values are preserved in the released repository.

### 3.7 Execution-grounded correctness

Two domains carry ground truth: forty arithmetic word problems, and forty "hard-verifiable" items — twenty-seven mathematical problems with a computed numeric answer and thirteen programming tasks with hidden assert-based tests. Mathematical responses end with a required "Final answer:" line, which is parsed and compared with tolerance; for code, the longest fenced block is extracted, the model's own asserts and print statements are removed, and the hidden tests are run in an isolated subprocess with a timeout. Two columns result: external correctness, and agreement with the reference response's own outcome on the same prompt. The second is the relevant one for a same-model comparison, and its floor is set by the second reference sample: at temperature 0.7 the unmodified model agrees with itself on only 31 of 40 hard-verifiable outcomes.

### 3.8 Reliability and validation

Twenty percent of items were judged twice by the primary judge: 86% exact agreement, 100% within one point, and a standard deviation of the difference of 0.37, so the judge-only standard deviation per rating is 0.26 — 8% of the variance of a paired score. Ten percent of items were also judged by claude-opus-5 (n = 176): Spearman ρ = 0.88 with the primary judge and a constant offset of −0.36. A random 29% were judged by gemini-3.1-pro (n = 226): ρ = 0.80, offset +0.02, per-arm offsets within ±0.2. Both second judge families place the arms in the same order as the primary. On the verifiable domains the judge's rating tracks the execution check: 6.07 when the checker finds the answer correct and 2.45 when wrong on hard-verifiable items, and 2.33 on the nine real arithmetic errors, with no item wrong by the checker rated 6 or above. We reviewed by hand a random sample of 32 judged responses, four per arm, and found no rating we would dispute. On the second target model the corresponding figures are 84% exact agreement, judge-only standard deviation 0.34 (12% of the paired-score variance), and ρ = 0.95 with claude-opus-5 (n = 140, offset −0.24).

### 3.9 What transfers

Nothing in Sections 3.2, 3.3, 3.6, or 3.8 depends on the intervention being an inference optimization. The dual-reference design, the exchangeability null, the equivalence bound and its power arithmetic, and the reliability checks apply unchanged to anything that alters a model's output distribution while intending to preserve its quality: context compression or summarization, KV-cache eviction, prompt compression, retrieval truncation, safety filtering, watermarking, distillation, or a changed system prompt. Two components need an intervention-specific analog. The implementation null (3.4) requires a setting of the intervention that is provably null; speculative decoding supplies one by theorem, a context compressor supplies one by returning the context unchanged, and a watermark supplies one when its bias on the sampling distribution is set to zero. The positive controls (3.5) require a prior measurement of a known loss, which a first study of a new intervention will not have and must bootstrap from pilots, as this one did. The execution check (3.7) applies wherever the prompt set contains verifiable items.

## 4. Experimental Setup

**Target models.** Qwen2.5-7B-Instruct (Alibaba) is the primary target; Llama-3.1-8B-Instruct (Meta) is the replication target ("Instruct" denotes the chat-tuned, instruction-following variant of each model, as opposed to the base pretrained model). Both run in bf16 as the reference. All arms for a model were generated on the same GPU (an RTX 5090), every 4-bit arm used the same quantization library, and every 3-bit arm used the same one, so that no difference between arms can arise from hardware or from a different quantizer implementation.

**Arms.** Table 1 lists the eight generations per prompt on Qwen. Llama has six: the two early-exit arms require a cache-repair map fitted to the model's own activations, which we fitted for Qwen only. NF4 is the bitsandbytes 4-bit NormalFloat format with double quantization (Dettmers et al., 2023); HQQ 3-bit is a data-free quantizer with group size 64 (Badri and Shaji, 2023). The 3-bit results below therefore characterize this quantizer, not the 3-bit weight budget in general: calibrated methods such as AWQ or GPTQ choose which channels to protect from activation statistics and may produce a different quality surface at the same bit width. Measuring that is an application of the protocol, not a limitation of it. The speculative arms use the NF4 model as drafter and bf16 as verifier, with eight drafted tokens per window; $\lambda$ is the leniency of the acceptance rule, $\min(1, r/\lambda)$ in place of $\min(1, r)$, so that $\lambda = 1$ is the strict, lossless rule and $\lambda = 0.2$ accepts almost every drafted token. $\lambda$ is unrelated to temperature: temperature shapes the model's next-token distribution (all arms use 0.7), whereas $\lambda$ relaxes the verifier's accept-or-reject decision on tokens the drafter proposes. The early-exit arms run the first 15 of 28 layers for every generated token and reconstruct the remaining layers' key and value cache entries from the layer-15 hidden state with a linear map fitted to the model's own activations (Kaplan, 2026); the refill variant recomputes the last 16 entries exactly every 16 tokens.

**Prompts and domains.** Five domains, 220 prompts: English explanatory prose (40), short Python functions (60), arithmetic word problems (40), Chinese explanatory prose (40), and hard-verifiable problems (40: 27 mathematical with computed answers, 13 programming tasks with hidden tests). The code domain (the 60 Python-function prompts) has more prompts than the others because its per-item variance is twice theirs. Prompts are released with the code.

**Generation.** Temperature 0.7 with no top-p or top-k truncation, one sample per arm per prompt, deterministic seeds, and a cap of 1,500 tokens (3,000 for Llama, which writes longer); responses that reach the cap are recorded as non-terminating. Speculative windows record the acceptance count and the expected acceptance rate (EAR) from the two distributions.

**Judging.** claude-sonnet-5 rated every response (1,760 items for Qwen, 1,320 for Llama), with 20% re-rated by the same model and 10% by claude-opus-5; a random 29% of the Qwen items were also rated by gemini-3.1-pro.

## 5. Results

### 5.1 The quality surface (Qwen2.5-7B)

Table 1 gives the pooled result and Table 2 the per-domain breakdown; Figure 1 shows both. Four arms are equivalent to the full model at the ±0.3 bound: the exchangeability null, the implementation null, lenient speculation, and NF4 quantization. NF4's pooled score is +0.01 [−0.10, +0.12], and every domain cell lies within a quarter of a point of zero, including Chinese (−0.16 [−0.42, +0.10]). Three arms are significantly worse. HQQ 3-bit loses 0.70 points pooled, and the loss is shaped: 0.09 on arithmetic, 0.46 on prose, 0.84 on code, 0.91 on Chinese, 1.11 on hard-verifiable problems. Early exit at layer 15 with refill loses 1.07 pooled and 2.54 on hard-verifiable problems; without refill, 1.67 and 3.29.

**Table 1. Pooled quality deltas, Qwen2.5-7B (n = 220 per arm). Verdict is the two-one-sided-tests decision at ±0.3.**

| Arm | Δ | 95% CI | Verdict | s.d. | Derailed | Mean EAR |
|---|---|---|---|---|---|---|
| reference sample 2 (bf16, null) | -0.05 | [-0.18, +0.08] | equivalent | 0.94 | 0.0% | — |
| speculative λ=1.0 (null) | +0.05 | [-0.06, +0.16] | equivalent | 0.84 | 0.9% | 0.950 |
| speculative λ=0.2 | -0.02 | [-0.14, +0.10] | equivalent | 0.92 | 0.9% | 0.952 |
| NF4 4-bit | +0.01 | [-0.10, +0.12] | equivalent | 0.84 | 1.8% | — |
| HQQ 3-bit | -0.70 | [-0.85, -0.55] | worse | 1.14 | 4.1% | — |
| exit d15, refill 16 | -1.07 | [-1.26, -0.87] | worse | 1.49 | 10.9% | — |
| exit d15, no refill | -1.67 | [-1.90, -1.43] | worse | 1.76 | 18.6% | — |

*EAR is recorded only in the speculative arms. The two reference samples had 0.0% derailed.*

**Table 2. Quality deltas by domain, Qwen2.5-7B (n = 40 per cell; 60 for code). ≡ equivalent at ±0.3; ↓ significantly worse; unmarked cells are undetermined at this n.**

| Arm | prose | code | arith | zh | hard |
|---|---|---|---|---|---|
| reference sample 2 (bf16, null) | +0.07 [-0.21, +0.36] | +0.08 [-0.20, +0.36] | -0.03 [-0.08, +0.03] ≡ | -0.12 [-0.40, +0.15] | -0.33 [-0.72, +0.07] |
| speculative λ=1.0 (null) | +0.16 [-0.05, +0.37] | +0.06 [-0.10, +0.22] ≡ | +0.01 [-0.01, +0.04] ≡ | -0.14 [-0.42, +0.14] | +0.16 [-0.30, +0.63] |
| speculative λ=0.2 | -0.09 [-0.27, +0.09] ≡ | +0.04 [-0.17, +0.25] ≡ | +0.01 [-0.01, +0.04] ≡ | -0.31 [-0.58, -0.04] ↓ | +0.21 [-0.30, +0.73] |
| NF4 4-bit | +0.09 [-0.12, +0.30] ≡ | -0.06 [-0.31, +0.19] ≡ | -0.04 [-0.14, +0.07] ≡ | -0.16 [-0.42, +0.10] | +0.24 [-0.12, +0.59] |
| HQQ 3-bit | -0.46 [-0.78, -0.15] ↓ | -0.84 [-1.17, -0.52] ↓ | -0.09 [-0.29, +0.12] ≡ | -0.91 [-1.24, -0.58] ↓ | -1.11 [-1.54, -0.69] ↓ |
| exit d15, refill 16 | -0.69 [-1.01, -0.36] ↓ | -1.02 [-1.45, -0.60] ↓ | -0.44 [-0.79, -0.08] ↓ | -0.66 [-0.95, -0.38] ↓ | -2.54 [-3.04, -2.04] ↓ |
| exit d15, no refill | -1.16 [-1.47, -0.85] ↓ | -1.62 [-2.10, -1.15] ↓ | -1.24 [-1.88, -0.60] ↓ | -1.04 [-1.34, -0.74] ↓ | -3.29 [-3.84, -2.74] ↓ |

*Arithmetic is at ceiling (every reference rated 7), so its judged cells carry little information; see Table 3.*

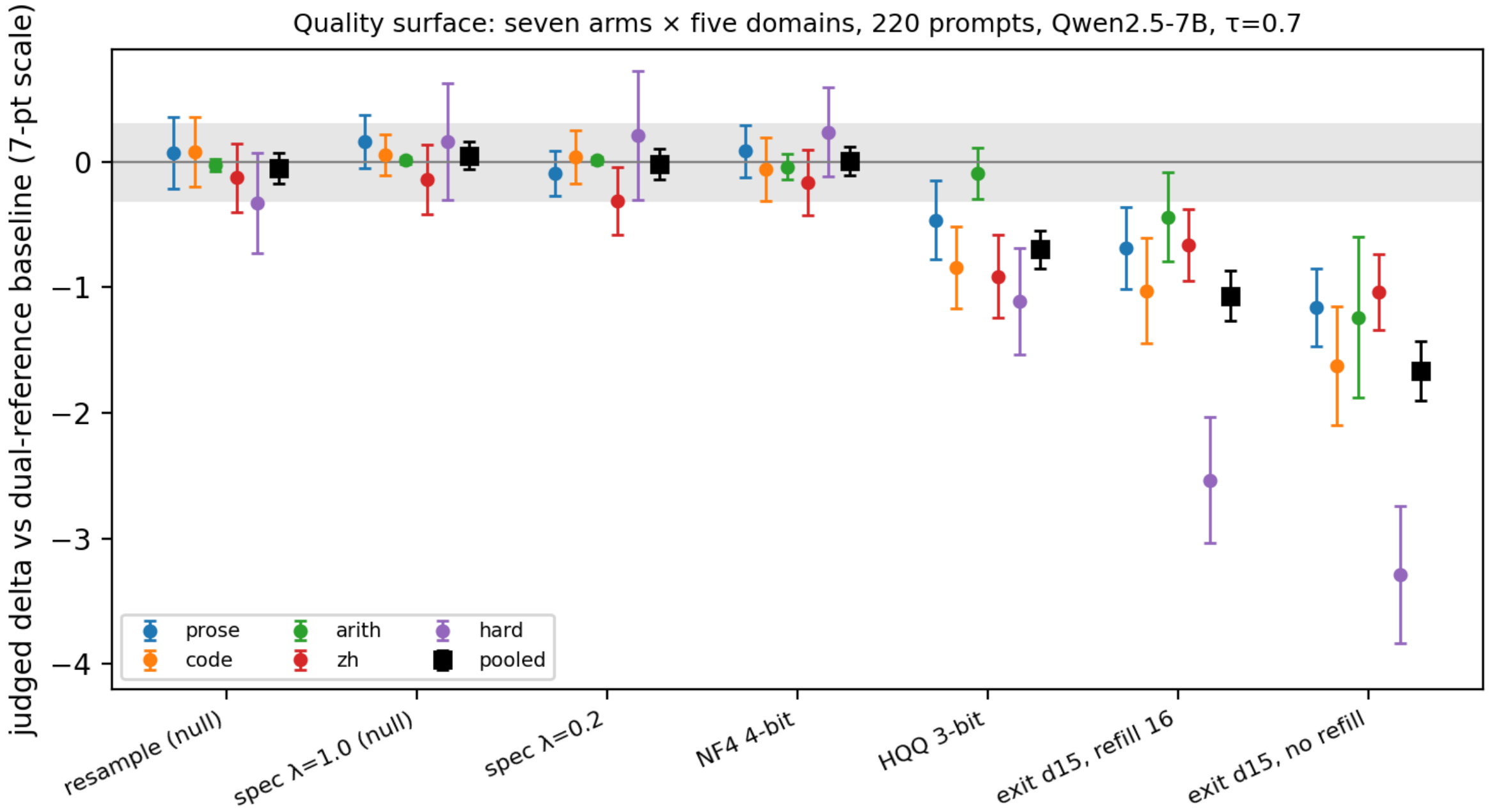

*Figure 1. Judged delta with 95% confidence interval for each arm, by domain and pooled, Qwen2.5-7B. The grey band is the ±0.3 equivalence bound.*

## 5.2 Execution-grounded correctness

Table 3 shows the checker's results. The reference solves 19 of 27 mathematical problems and passes all tests on 8 of 13 programs; the second reference sample, 15 and 7. Quantization to 4 bits and both speculative arms stay within three items of the reference on both counts. HQQ 3-bit drops to 12 and 2. Both early-exit arms collapse: 6 and 4 of 27 in mathematics, 0 of 13 in code, and exact refill every 16 tokens does not rescue them. The agreement column — whether an arm reaches the same outcome as the reference on the same prompt — has its floor set by the second reference sample at 31 of 40: the unmodified model disagrees with itself on 22% of hard outcomes at temperature 0.7, which is the noise any outcome-flip statistic has to clear. Against that floor, only the early-exit arms (19 and 15) are clearly below.

**Table 3. Execution-grounded correctness, Qwen2.5-7B.**

| Arm | Math correct | Code: all tests pass | Code tests passed | Arith: same outcome as ref. | Hard: same outcome as ref. |
|---|---|---|---|---|---|
| reference sample 1 (bf16) | 19/27 | 8/13 | 30/48 | — | — |
| reference sample 2 (bf16, null) | 15/27 | 7/13 | 30/48 | 40/40 | 31/40 |
| speculative λ=1.0 (null) | 18/27 | 8/13 | 29/48 | 40/40 | 31/40 |
| speculative λ=0.2 | 18/27 | 9/13 | 37/48 | 40/40 | 30/40 |
| NF4 4-bit | 18/27 | 6/13 | 27/48 | 40/40 | 35/40 |
| HQQ 3-bit | 12/27 | 2/13 | 16/48 | 39/40 | 27/40 |
| exit d15, refill 16 | 6/27 | 0/13 | 15/48 | 40/40 | 19/40 |
| exit d15, no refill | 4/27 | 0/13 | 7/48 | 32/40 | 15/40 |

*Arithmetic outcomes: every arm except HQQ 3-bit (39/40) and early exit without refill (32/40) matches the reference on all 40; the judge rated all nine real arithmetic errors between 1 and 4.*

### 5.3 Replication on Llama-3.1-8B

The same prompts, seeds, quantizers, and judge on Llama-3.1-8B-Instruct (Tables 4 and 5, Figure 2) reproduce the ordering and change the magnitudes. NF4 is equivalent at the bound pooled (−0.10 [−0.24, +0.04]) and inside it in every domain. Lenient speculation is equivalent. HQQ 3-bit is far worse on Llama than on Qwen: −1.84 [−2.08, −1.60] pooled, with code the worst domain (−3.11) rather than hard-verifiable problems, Chinese (−2.02) again worse than English prose (−1.49), and arithmetic again untouched (−0.09). Twenty-three percent of Llama's 3-bit responses fall into repetition, against 1% for Qwen's. The same quantizer, at the same bit width, costs one model 0.7 points and the other 1.8, and damages them in different places.

The implementation null on Llama reads −0.14 [−0.27, −0.01]: inside the equivalence bound, but significantly below zero. The deficit is concentrated in the hard domain (−0.46 [−0.81, −0.11]), is reproduced by the lenient arm (−0.44), is absent when the NF4 drafter is sampled on its own (−0.01), survives the exclusion of every non-terminating response (−0.45 on the 28 clean prompts), and is corroborated mechanically: hard responses from the speculative loop average 1,185 tokens against 874 for the reference. On Qwen the same code reads +0.05. A control arm in which the drafter is the unmodified model itself under the strict rule — so that its output should match the unmodified model exactly — reads −0.16 [−0.56, +0.24] on the same hard prompts, which neither confirms nor excludes batched-arithmetic numerics as the cause; the residual −0.29 between control and drafted arms is not significant at $n = 40$. The reading is a property of this implementation on this model, not of speculative decoding, whose strict rejection rule is exact by construction. We report it as what the null is for: a model-specific effect below the bound that no surface statistic would have flagged, detected and bounded but not explained at this sample size.

**Table 4. Pooled quality deltas, Llama-3.1-8B (n = 220 per arm).**

| Arm | Δ | 95% CI | Verdict | s.d. | Derailed | Repetitive |
|---|---|---|---|---|---|---|
| reference sample 2 (bf16, null) | +0.09 | [-0.06, +0.25] | equivalent | 1.17 | 7.3% | 7% |
| speculative λ=1.0 (null) | -0.14 | [-0.27, -0.01] | equivalent | 0.97 | 9.1% | 11% |
| speculative λ=0.2 | -0.10 | [-0.23, +0.03] | equivalent | 0.98 | 8.2% | 8% |
| NF4 4-bit | -0.10 | [-0.24, +0.04] | equivalent | 1.08 | 8.2% | 9% |
| HQQ 3-bit | -1.84 | [-2.08, -1.60] | worse | 1.81 | 36.4% | 23% |

*Repetitive: fraction of responses with more than 10% repeated 8-grams. Reference: 7%.*

**Table 5. Quality deltas by domain, Llama-3.1-8B.**

| Arm | prose | code | arith | zh | hard |
|---|---|---|---|---|---|
| reference sample 2 (bf16, null) | +0.33 [-0.00, +0.65] | +0.18 [-0.13, +0.49] | +0.12 [-0.21, +0.46] | -0.15 [-0.49, +0.19] | -0.07 [-0.54, +0.39] |
| speculative λ=1.0 (null) | -0.01 [-0.31, +0.28] ≡ | -0.09 [-0.38, +0.20] | +0.04 [-0.16, +0.23] ≡ | -0.20 [-0.49, +0.09] | -0.46 [-0.81, -0.11] ↓ |
| speculative λ=0.2 | +0.11 [-0.12, +0.34] | -0.07 [-0.38, +0.23] | +0.09 [-0.11, +0.28] ≡ | -0.20 [-0.49, +0.09] | -0.44 [-0.81, -0.07] ↓ |
| NF4 4-bit | -0.11 [-0.34, +0.12] | -0.17 [-0.49, +0.14] | -0.04 [-0.29, +0.22] ≡ | -0.12 [-0.47, +0.22] | -0.01 [-0.46, +0.44] |
| HQQ 3-bit | -1.49 [-1.89, -1.09] ↓ | -3.11 [-3.57, -2.64] ↓ | -0.09 [-0.39, +0.22] | -2.02 [-2.47, -1.58] ↓ | -1.86 [-2.44, -1.28] ↓ |

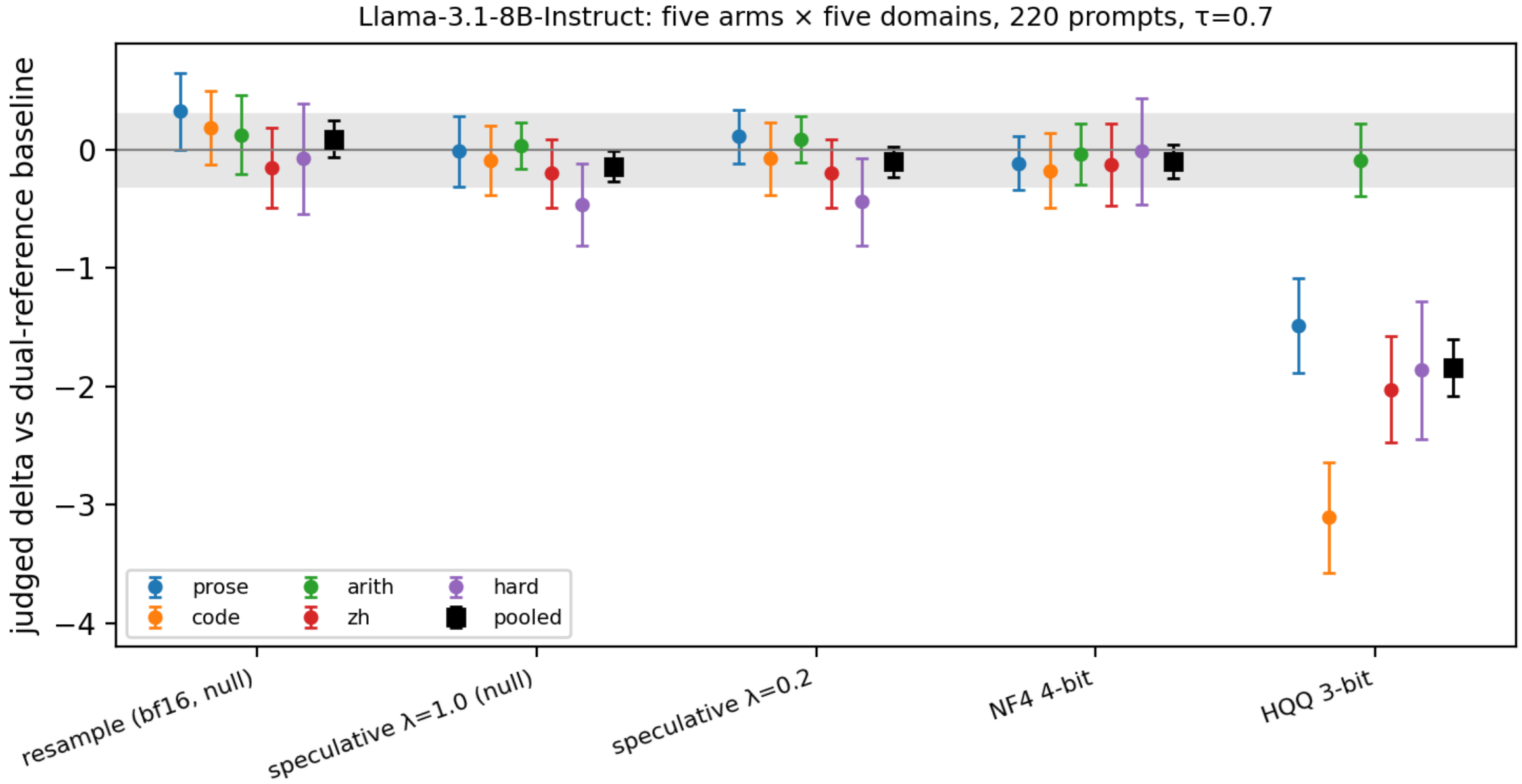

*Figure 2. Judged delta with 95% confidence interval, Llama-3.1-8B.*

Llama's external correctness column is thin — the bf16 model solves 6 of 27 mathematical problems and passes 5 of 13 programs — so the comparable column across models is agreement with the reference's outcome: reference sample 2, 36/40, NF4 34, both speculative arms 36–37, HQQ 3-bit 31 (and 0 of 27 externally correct, 5 of 48 code tests).

### 5.4 The instrument, both models

**Table 6. Instrument characteristics.**

| Quantity | Qwen2.5-7B | Llama-3.1-8B |
|---|---|---|
| Same text judged twice: exact / within 1 point | 86% / 100% | 84% / 99% |
| Judge-only s.d. per rating | 0.26 | 0.34 |
| Judge share of paired-score variance | 8% | 12% |
| Prompt-effect ICC (dual reference) | 0.12 | 0.09 |
| Reference mean rating / at ceiling | 6.33 / 68% | 5.46 / 42% |
| Ceiling by domain (prose, code, arith, zh, hard) | 65%, 78%, 100%, 50%, 40% | 33%, 58%, 95%, 0%, 15% |
| Paired-score s.d. by domain | 0.95, 1.41, 1.07, 0.96, 2.03 | 1.13, 1.87, 0.75, 1.35, 1.55 |
| Second judge, claude-opus-5: ρ / offset | 0.88 / −0.36 (n = 176) | 0.95 / -0.24 (n = 140) |
| Third judge, gemini-3.1-pro: ρ / offset | 0.80 / +0.02 (n = 226) | — |
| Rating when checker says correct / wrong (hard) | 6.07 / 2.45 | 5.84 / 2.04 |
| Items wrong by checker yet rated ≥ 6 | 0 | 0 |

## 6. What the Instrument Shows That Proxies Do Not

**The noise is the model's, not the judge's.** Re-rating the same text moves the score by 0.26 points on Qwen and 0.34 on Llama; a fresh sample of the same model on the same prompt moves it by 0.94 and 1.17. The judge is 8–12% of the variance of a paired comparison. Better judges will not shrink the intervals; more prompts will, and the arithmetic in Section 3.6 says how many. Studies that compare compressed models on tens of prompts, or on one sample per prompt without a second reference draw, risk reporting sampling noise as effects.

**Ceiling.** Qwen's reference is rated 7 on 68% of prompts and on every arithmetic prompt; Llama's on 42% and 95%. Above the ceiling an arm can only be seen to lose, and arithmetic cannot be judged at all — every judged arithmetic cell in Tables 2 and 5 is zero. The execution column is what measures that domain, and it finds the early-exit errors (8 of 40 without refill) and the one arithmetic problem the 3-bit model got wrong, none of which the ratings could register. When judged results are pinned at the maximum value, measurement requires another technique, and objective verifiability serves that purpose in the domains that permit it.

**Using distribution fidelity as a measure of quality impact is inadequate: the actual damage varies widely by domain and by model.** The expected acceptance rate — one minus the total-variation distance between the compressed and full models' next-token distributions, and the quantity speculative decoding optimizes — was measured on the 3-bit model's own responses. It ranks the domains by how often the two models disagree — a lower EAR means more disagreement — and disagreement is highest in open-ended prose (EAR 0.84 on Qwen) and lowest in arithmetic (0.95); Chinese is lower still, at 0.81. Judged damage ranks them by consequence: arithmetic 0.09, prose 0.46, code 0.84, Chinese 0.91, hard 1.11 on Qwen. Prose, where the models disagree most, loses least among the non-arithmetic domains; hard problems, where they disagree least, lose most. On Llama the same inversion appears with code (EAR 0.86, loss 3.11). At the level of individual responses, a response's EAR is a weak predictor of its judged loss: for the 3-bit arm, Spearman ρ = 0.19 on Qwen and 0.25 on Llama over 220

responses, at most 0.3 within any domain, and near zero or negative in the domains that lose most (−0.02 on Qwen's hard problems; −0.42 on Llama's Chinese, where high local agreement often marks a repetition loop). Sorting the 220 responses by EAR shows where the weak correlation comes from: the quarter with the highest fidelity loses noticeably less, and the other three quarters lose about the same, so fidelity discriminates only at the top. On Qwen the highest quarter loses 0.46 points against 0.73 to 0.85 for the other three; on Llama, 1.01 against 1.98 to 2.19. Across domains the five-domain means correlate at 0.3 and 0.5. EAR identifies arithmetic as safe and Chinese as at risk in both models, and misranks the rest, because when the models disagree on a prose token the alternatives are usually interchangeable, whereas in a proof the alternative is usually a wrong number. EAR measured on the 3-bit model's own responses equals EAR measured on the reference responses to three decimals on Qwen: the 3-bit model does not drift further from the full model as its own output accumulates; the damage is in which tokens it chooses.

**Certainty is not consequence.** The same mismatch between certainty and consequence holds at the token level. In an earlier campaign on Qwen2.5-7B, digits and punctuation were emitted with nearly identical confidence (0.999 and 0.989), yet substituting a plausible alternative changed the judged outcome in 23.3% of digit positions and 3.3% of punctuation positions ($n = 354$). An acceptance rule keyed on confidence — the basis of every relaxed speculative-decoding scheme and of confidence-thresholded early exit — treats those positions alike. The early-exit arms in Table 3 show this same lack of correlation: the answers may contain fluent prose, but still have wrong numbers. The same distinction has begun to appear on the mechanism side: pivot-aware acceptance (Ziashahabi et al., 2025) rejects only tokens predicted to change the final outcome, which presupposes that consequence and probability are different quantities — the claim our token-level measurement supports.

**Perplexity is not a suitable proxy for quality.** Early-exit configurations in the pilot occasionally produced responses that degenerated into looping nonsense. One such response had lower perplexity (1.151) than the reference (1.239), because repetition is easy to predict; and token-level fidelity and length-based proxies rated as acceptable 37% of the responses the judge found unacceptable. Non-termination is the cleanest example: the three Qwen responses that never terminated were rated 1.67 on average, and 23% of Llama's 3-bit responses fall into repetition, which perplexity rewards.

**The judge sees consequences; cheaper signals do not.** We expected the judges' ratings to behave like the proxies and under-read the early-exit collapse; it did not, and no item the objective-answer checker marked wrong was rated 6 or above on either model. The judge is therefore a valid instrument. The practical question for inference-time gating is whether any cheap signal — perplexity, confidence, agreement, EAR — tracks quality well enough to stand in for it, and the answer from this study is clear: none of these measures are adequate, and should be reassessed and used with caution by the LLM optimization community.

**Calibration is not a cause of the measured differences by domain.** The 3-bit quantizer used here is data-free, so its domain dependence cannot be a calibration artifact, and Chimoto et al. (2026) show on the same model that language-matched AWQ calibration moves Chinese perplexity by only 1–3% and leaves unchanged which weight channels the quantizer protects at higher precision, and Sun et al. (2024) show that the largest activation values occur in the same few dimensions for every input. We tentatively attribute the twofold spread between English and Chinese damage at 3 bits, and the fourfold spread between arithmetic and code on Llama at 3 bits, to the bit budget and model capacity.

## 7. Limitations

The primary judge is one model family; the two secondary families agree on ordering and differ by a constant offset, which is why every comparison is within one judge, but a judge with a different rubric or a pairwise protocol might resolve differently. Two target models of 7–8 billion parameters were measured; larger models tolerate low bit widths better and the magnitudes will not transfer, though the method does. Five domains and 220 prompts give ±0.11 pooled and ±0.3 per domain; per-domain equivalence claims at this resolution need 80–150 prompts per domain, and the hard-verifiable domain, with a paired standard deviation of 2.0, needs the most. Early exit was tested at one depth (layer 15) and cache refill at one interval (every 16 tokens); the trade-off across depths and intervals was not mapped. On Llama the lossless speculative arm read −0.14, below zero but within the bound; we localized the deficit to long multi-step responses but did not determine its cause. Finally, the execution-grounded column exists only where ground truth exists; for prose and Chinese, the judge is the only instrument, and its validity there rests on the reliability figures and the second families rather than on an external objective check.

## 8. Conclusion

An LLM judge, used with two null controls, positive controls, a dual-reference paired design, a pre-specified equivalence bound, and an execution check where one is possible, can be used to measure the effect of lossy inference optimizations on output quality to a stated resolution, on any model and any domain, and lets techniques from different families be placed on one scale.

In our tests on two 7–8 billion-parameter models, we found 4-bit NF4 and lossless speculation indistinguishable from full precision at 0.3 rating points in every test domain. In contrast, we found that 3-bit quantization reduced judged quality in a domain- and model-dependent way. Similarly, we found that early exit with cache repair severely impacted judged quality on multi-step problems, yet was nearly harmless when tested on prose. The proxies the literature relies on — perplexity, token agreement, distribution fidelity, confidence — rank neither the domains nor the arms the way our quality judgment outcomes do, because they substitute technical details of the model performance for a proper evaluation of overall output quality.

Our protocol, prompts, rubric, and analysis are publicly available at https://github.com/jerrykaplan/Calibrated-Instrument for reproducibility and as a tool for others to use. Nothing in the protocol is specific to inference optimizations: it applies to any intervention that changes a model's outputs while intending to preserve their quality.

## 9. Acknowledgments and Declaration of Generative AI and AI-assisted Technologies in the Manuscript Preparation Process

Experiments, analysis, and drafting were carried out in collaboration with Claude Fable 5.1 and Claude Opus 5 (Anthropic), AI systems operating under the author's direction across an extended series of working sessions. All experimental designs were approved and/or originated by the author; all claims were verified against archived experimental records. Responsibility for all content of the published article rests solely with the author.

## References

Bachmann, G., Anagnostidis, S., Pumarola, A., Georgopoulos, M., Sanakoyeu, A., Du, Y., Schönfeld, E., Thabet, A., and Kohler, J. (2025). Judge Decoding: Faster speculative sampling requires going beyond model alignment. International Conference on Learning Representations (ICLR 2025). arXiv:2501.19309.

Badri, H. and Shaji, A. (2023). Half-Quadratic Quantization of large machine learning models. Mobius Labs technical report. https://mobiusml.github.io/hqq_blog/

Borgersen, K. A. K. and Goodwin, M. (2025). English k-quantization of LLMs does not disproportionately diminish multilingual performance. arXiv:2503.03592.

Chen, C., Borgeaud, S., Irving, G., Lespiau, J.-B., Sifre, L., and Jumper, J. (2023). Accelerating large language model decoding with speculative sampling. arXiv:2302.01318.

Chimoto, E. A., Elhoushi, M., and Bassett, B. (2026). Calibrating beyond English: Language diversity for better quantized multilingual LLMs. Proceedings of the 19th Conference of the European Chapter of the Association for Computational Linguistics (EACL 2026), pages 4822–4838.

Dettmers, T., Pagnoni, A., Holtzman, A., and Zettlemoyer, L. (2023). QLoRA: Efficient finetuning of quantized LLMs. Advances in Neural Information Processing Systems 36 (NeurIPS 2023).

Dutta, A., Krishnan, S., Kwatra, N., and Ramjee, R. (2024). Accuracy is not all you need. Advances in Neural Information Processing Systems 37 (NeurIPS 2024). arXiv:2407.09141.

Elhoushi, M., Shrivastava, A., Liskovich, D., Hosmer, B., Wasti, B., Lai, L., Mahmoud, A., Acun, B., Agarwal, S., Roman, A., Aly, A., Chen, B., and Wu, C.-J. (2024). LayerSkip: Enabling early exit inference and self-speculative decoding. Proceedings of the 62nd Annual Meeting of the Association for Computational Linguistics (ACL 2024).

Gao, I., Liang, P., and Guestrin, C. (2024). Model equality testing: Which model is this API serving? arXiv:2410.20247.

Garipov, R., Shchegoleva, A., Belova, A., and Kuznedelev, D. (2025). AutoJudge: Judge decoding without manual annotation. arXiv:2504.20039.

Georganas, E., Kalamkar, D., Kozlov, A., and Heinecke, A. (2025). ML-SpecQD: Multi-level speculative decoding with quantized drafts. arXiv:2503.13565.

Helcig, F., Kurtić, E., and Alistarh, D. (2026). Statistically-lossless quantization of large language models. Conference on Language Modeling (COLM 2026). arXiv:2605.02404.

Jaiswal, A., Gan, Z., Du, X., Zhang, B., Wang, Z., and Yang, Y. (2024). Compressing LLMs: The truth is rarely pure and never simple. International Conference on Learning Representations (ICLR 2024). arXiv:2310.01382.

Kaplan, J. (2026). One size does not fit all: Setting inference depth from the questions a deployment actually asks. Manuscript (arXiv:2609.14144).

Kübler, J. M., Budhathoki, K., Kleindessner, M., Zhou, X., Yin, J., Khetan, A., and Karypis, G. (2026). When LLMs get significantly worse: A statistical approach to detect model degradations. International Conference on Learning Representations (ICLR 2026). arXiv:2602.10144.

Kurtić, E., Marques, A., Pandit, S., Kurtz, M., and Alistarh, D. (2025). "Give me BF16 or give me death"? Accuracy–performance trade-offs in LLM quantization. Proceedings of the 63rd Annual Meeting of the Association for Computational Linguistics (ACL 2025). arXiv:2411.02355.

Lakens, D. (2017). Equivalence tests: A practical primer for t tests, correlations, and meta-analyses. Social Psychological and Personality Science, 8(4), 355–362.

Leviathan, Y., Kalman, M., and Matias, Y. (2023). Fast inference from transformers via speculative decoding. Proceedings of the 40th International Conference on Machine Learning (ICML 2023).

Marchisio, K., Dash, S., Chen, H., Aumiller, D., Üstün, A., Hooker, S., and Ruder, S. (2024). How does quantization affect multilingual LLMs? Findings of the Association for Computational Linguistics: EMNLP 2024. arXiv:2407.03211.

Miller, E. (2024). Adding error bars to evals: A statistical approach to language model evaluations. arXiv:2411.00640.

Schuster, T., Fisch, A., Gupta, J., Dehghani, M., Bahri, D., Tran, V. Q., Tay, Y., and Metzler, D. (2022). Confident adaptive language modeling. Advances in Neural Information Processing Systems 35 (NeurIPS 2022).

Sun, M., Chen, X., Kolter, J. Z., and Liu, Z. (2024). Massive activations in large language models. Conference on Language Modeling (COLM 2024). arXiv:2402.17762.

Tiwari, R., Xi, H., Tomar, A., Hooper, C., Kim, S., Horton, M., Najibi, M., Mahoney, M. W., Keutzer, K., and Gholami, A. (2025). QuantSpec: Self-speculative decoding with hierarchical quantized KV cache. arXiv:2502.10424.

Usami, H., Hara, K., Tsuboi, A., and Matsuda, N. (2026). LLM judges have dark current: A psychometric datasheet for LLM-as-a-judge evaluation. arXiv:2606.15610.

Xia, R., Ribar, L., and Balanca, P. (2026). A practical investigation of training-free relaxed speculative decoding. arXiv:2607.08690.

Yagubyan, A. (2026). The coin flip judge? Reliability and bias in LLM-as-a-judge evaluation. arXiv:2606.13685.

Zhao, J., Lu, W., Wang, S., Kong, L., and Wu, C. (2025). QSpec: Speculative decoding with complementary quantization schemes. Proceedings of the 2025 Conference on Empirical Methods in Natural Language Processing (EMNLP 2025). arXiv:2410.11305.

Ziashahabi, A., Bakman, Y. F., Yaldiz, D. N., El-Khamy, M., Karimireddy, S. P., and Avestimehr, S. (2025). Reject Only Critical Tokens: Pivot-Aware Speculative Decoding. The First Workshop on Efficient Reasoning, NeurIPS 2025. arXiv:2511.00351.

## Appendix A. Judge rubric

*The system prompt given to the judge, verbatim. Each call supplied the prompt and one response inside <request> and <response> tags, and nothing else.*

```
You are rating the quality of a response produced by a language model.

Rate how well the response answers the request, on this scale:

7  Excellent. Complete, correct, well-formed. Nothing a reader would object to.
6  Good. Minor awkwardness or a small omission, but fully usable.
5  Acceptable. Noticeable rough edges: slight redundancy, a clumsy sentence, a
   small factual slip, or an ending that trails rather than concludes.
4  Mediocre. Real problems: repeats itself, drifts off the question, contains a
   clear factual or logical error, or is visibly incomplete.
3  Poor. Substantially fails the request, though still recognisable as an
   attempt at it.
2  Bad. Largely incoherent, or degenerates part-way through.
1  Broken. Degenerate repetition, no meaningful content, or unusable output.

Also decide, separately: did the response DERAIL? Derailed means it lost the
thread -- looping on a word or phrase, abandoning the topic, producing garbled
or unrelated text, or running on without ever concluding. A response can be
mediocre without derailing.

And decide: did the response RECOVER? If the response contains a damaged,
garbled, or confused stretch but returns to coherent, on-task output afterward
and concludes properly, recovered is true. If it never goes wrong at all,
recovered is true. If it goes wrong and stays wrong, recovered is false.

If the response contains code, judge whether the code is syntactically valid
and does what was asked. Unbalanced brackets, undefined names, or truncated
blocks are serious faults.

Respond with JSON only, no other text, no markdown fences:
{"rating": <1-7>, "derailed": <true|false>, "recovered": <true|false>, "reason": "<at
most 15 words>"}
```

## Appendix B. Prompt sets

Four standard domains use the prompt sets released with Kaplan (2026): 60 English explanatory prompts ("Explain why the seasons change"), 60 short Python function completions ("Complete this Python function: def reverse_words(s): …"), 60 arithmetic word problems ("A pool fills at 65 liters per minute for 11 minutes. How much water? Show your work."), and 60 Chinese explanatory prompts ("请解释四季变化的原因。"). This study used the first 40 of each (all 60 for code). The hard-verifiable set was written for this study; it follows in full, with the ground truth used by the checker. Mathematical prompts were suffixed with "Show your reasoning, then give the final answer on its own last line in the form 'Final answer: <value>'." and programming prompts with "Write clean Python with the exact function/class name requested and include at least three assert-based test cases." The hidden tests are in the repository.

| ID | Kind | Prompt | Truth / tests |
|---|---|---|---|
| H01 | math | Let S be the sum of all positive integers n < 1000 such that n^2 + n + 41 is divisible by 43. Find S. | 23714 |
| H02 | math | A rope bridge can hold at most 2 people. Four people cross at night with one flashlight; they take 1, 2, 5, and 10 minutes respectively, and a pair walks at the slower person's pace. Find the minimum total crossing time in minutes and give a schedule that achieves it. | 17 |
| H03 | math | Compute the exact value of the integral from 0 to 1 of ln(x) * ln(1-x) dx, showing the series manipulation, and give its numerical value to four decimal places. | 0.3551 (±0.001) |
| H04 | math | 41 people stand in a circle numbered 1 to 41. Starting the count at person 1, every third person is eliminated (persons 3, 6, 9, ... are removed first) and the count continues around the shrinking circle. Which numbered person is the last one remaining? | 31 |
| H05 | math | You have two identical eggs and a 100-floor building. An egg breaks if dropped from floor F or above and survives below F. What is the minimum number of drops that guarantees finding F in the worst case? Prove the bound and give the strategy. | 14 |
| H06 | math | How many trailing zeros does 250! have when written in base 10? | 62 |
| H07 | math | What are the last two digits of 7^2026? | 49 |
| H08 | math | How many positive divisors does 75600 have? | 120 |
| H09 | math | What is the sum of the decimal digits of 2^100? Compute 2^100 explicitly and then sum its digits. | 115 |
| H10 | math | What is the smallest positive integer that has exactly 24 positive divisors? Justify minimality. | 360 |
| H11 | math | How many ordered pairs of positive integers (x, y) satisfy 1/x + 1/y = 1/12? | 15 |
| H12 | math | Four fair six-sided dice are rolled. What is the probability that the sum is exactly 14? Give the answer as a reduced fraction. | 73/648 |
| H13 | math | A fair six-sided die is rolled repeatedly until every face has appeared at least once. What is the expected number of rolls? Give the exact value and its decimal. | 14.7 (±0.01) |
| H14 | math | How many five-digit positive integers have digits that are strictly increasing from left to right? | 126 |

| H15 | math | Compute 2^2026 mod 1000. | 864 |
|---|---|---|---|
| H16 | math | In how many ways can a 2-by-12 rectangle be tiled with 1-by-2 dominoes? Derive the recurrence and evaluate it. | 233 |
| H17 | math | Evaluate the infinite series sum from n=1 to infinity of 1/(n(n+2)). Give the exact value. | 3/4 |
| H18 | math | How many permutations of 7 distinct objects leave no object in its original position (derangements)? | 1854 |
| H19 | math | What is the sum of all three-digit palindromic numbers (such as 121 or 909)? | 49500 |
| H20 | math | Compute the sum 1^3 + 2^3 + 3^3 + ... + 100^3. | 25502500 |
| H21 | math | Factor 2^32 + 1 = 4294967297 into primes and state its largest prime factor. | 6700417 |
| H22 | math | A triangle has side lengths 13, 14, and 15. Find its area and the radius of its inscribed circle. State the inradius as the final answer. | 4 |
| H23 | math | How many distinct labeled trees are there on 6 labeled vertices? State and apply the relevant theorem. | 1296 |
| H24 | math | How many surjective (onto) functions are there from a set of 6 elements to a set of 3 elements? | 540 |
| H25 | math | How many prime numbers are less than 1000? | 168 |
| H26 | math | How many integers x with 0 <= x < 1000 satisfy x^2 ≡ 1 (mod 1000)? List them. | 8 |
| H27 | math | Compute the binomial coefficient C(30, 15) and state its last three digits. | 520 |
| H28 | code | Implement an LRU cache in Python as a class LRUCache(capacity) with methods get(key) -> value or -1, and put(key, value), both O(1), using a doubly linked list and a dict, without using OrderedDict or functools.lru_cache. Include a worked trace demonstrating eviction order. | 1 test |
| H29 | code | Write a Python function match(pattern, text) -> bool that implements a simplified regular-expression language supporting literal characters, '.', postfix '*', and '\|' with correct precedence (concatenation binds tighter than '\|'), using a recursive-descent parser over the pattern and a matcher that requires the entire text to match. Include edge-case tests. | 6 tests |
| H30 | code | Write a Python generator all_trees(n) that yields every structurally distinct binary tree with n nodes exactly once, in a canonical order, and a function count_trees(n) that counts them | 3 tests |
| | | without enumeration. Show that the counts match the Catalan numbers for n up to 6 and print count_trees(10). | |
| H31 | code | Write a Python function topo_sort(n, edges) that returns a topological ordering of the vertices 0..n-1 of the directed graph given by the edge list, or None if the graph contains a cycle. Use Kahn's algorithm or DFS and explain the cycle detection. | 3 tests |
| H32 | code | Write a Python function dijkstra(adj, src) that takes an adjacency dict mapping each vertex to a list of (neighbor, weight) pairs with non-negative weights and returns a dict of shortest-path distances from src to every reachable vertex, using a binary heap. Explain why the algorithm is correct with non-negative weights. | 2 tests |
| H33 | code | Write a Python function lis(seq) that returns a longest strictly increasing subsequence of a list of integers in O(n log n) time, reconstructing the actual subsequence (not just its length). Explain the patience-sorting idea. | 4 tests |
| H34 | code | Write a Python function is_balanced(code) that checks whether the brackets (), [], {} in a string of source code are properly balanced and nested, while ignoring any brackets that appear inside double-quoted string literals (assume no escaped quotes). | 5 tests |
| H35 | code | Write a Python function evaluate(expr) that evaluates an arithmetic expression string containing integers, + - * /, parentheses, and unary minus, with standard precedence and left associativity, without using eval or exec. Use a recursive-descent parser and return a float. | 4 tests |
| H36 | code | Write two Python functions: to_roman(n) for 1 <= n <= 3999, and from_roman(s) that converts a Roman numeral back to an integer and raises ValueError for malformed input such as 'IIII' or 'IM'. Explain the subtractive-notation rules you enforce. | 5 tests |
| H37 | code | Write a Python function merge_k(lists) that merges k sorted lists of integers into one sorted list in O(N log k) time using a heap, where N is the total number of elements. Handle empty input lists. | 4 tests |
| H38 | code | Write a Python function count_components(n, edges) that returns the number of connected components in an undirected | 3 tests |

| | | graph on vertices 0..n-1, using a union-find structure with path compression and union by rank. | |
|---|---|---|---|
| H39 | code | Write a Python function day_of_week(year, month, day) that returns the English weekday name for a Gregorian calendar date, implemented from first principles (Zeller's congruence or equivalent) without importing datetime or calendar. | 4 tests |
| H40 | code | Write a Python function longest_palindrome(s) that returns the longest palindromic substring of s (any one of them if there are ties) in O(n^2) time or better, and explain the expand-around-center approach. | 4 tests |

## Appendix C. Response lengths and generation time by arm

*Mean and median response length in tokens, mean wall-clock seconds per response on the RTX 5090, fraction of responses reaching the token cap (1,500 for Qwen, 3,000 for Llama), and fraction with more than 10% repeated 8-grams. The 3-bit arm's wall time reflects the HQQ library's unoptimized PyTorch dequantization kernel, not a property of 3-bit inference.*

### Qwen2.5-7B

| Arm | n | Mean tokens | Median tokens | Seconds | At cap | Repetitive |
|---|---|---|---|---|---|---|
| reference sample 1 | 220 | 395 | 369 | 4.9 | 0.0% | 1% |
| reference sample 2 | 220 | 395 | 354 | 4.9 | 0.0% | 1% |
| speculative λ=1.0 | 220 | 400 | 351 | 5.0 | 0.5% | 1% |
| speculative λ=0.2 | 220 | 401 | 378 | 4.2 | 0.0% | 1% |
| NF4 4-bit | 220 | 406 | 376 | 3.4 | 0.0% | 1% |
| HQQ 3-bit | 220 | 414 | 380 | 35.2 | 0.5% | 1% |
| exit d15, refill 16 | 220 | 407 | 372 | 5.6 | 0.0% | 2% |
| exit d15, no refill | 220 | 365 | 335 | 4.7 | 0.5% | 1% |

### Llama-3.1-8B

| Arm | n | Mean tokens | Median tokens | Seconds | At cap | Repetitive |
|---|---|---|---|---|---|---|
| reference sample 1 | 220 | 442 | 391 | 5.9 | 1.4% | 7% |
| reference sample 2 | 220 | 448 | 388 | 6.0 | 2.7% | 7% |
| speculative λ=1.0 | 220 | 505 | 398 | 7.7 | 4.1% | 11% |
| speculative λ=0.2 | 220 | 438 | 376 | 5.7 | 2.7% | 8% |
| NF4 4-bit | 220 | 438 | 384 | 4.6 | 1.8% | 9% |
| HQQ 3-bit | 220 | 495 | 346 | 44.2 | 4.1% | 23% |

## Appendix D. Third judge family (gemini-3.1-pro)

A random 29% of the Qwen items (226 shared with the primary judge after duplicate handling) were rated by gemini-3.1-pro with the same rubric. Overall: Spearman ρ = 0.80 with claude-sonnet-5, mean absolute difference 0.35, exact agreement 70%, within one point 96%, mean ratings 6.09 (Claude) and 6.11 (Gemini). One item of 226 differed by three or more points (a prose response Gemini faulted for a factual error about supply shifts that Claude rated 7). By arm:

| Arm | n | Mean, Claude | Mean, Gemini | Gemini − Claude | ρ |
|---|---|---|---|---|---|
| reference sample 1 | 30 | 6.3 | 6.37 | +0.07 | 0.7 |
| reference sample 2 | 25 | 6.64 | 6.6 | −0.04 | 0.62 |
| speculative λ=1.0 | 30 | 6.67 | 6.57 | −0.10 | 0.62 |
| speculative λ=0.2 | 30 | 6.4 | 6.3 | −0.10 | 0.85 |
| NF4 4-bit | 31 | 6.23 | 6.23 | 0.00 | 0.77 |
| HQQ 3-bit | 28 | 5.46 | 5.5 | +0.04 | 0.83 |
| exit d15, refill 16 | 25 | 5.64 | 5.84 | +0.20 | 0.89 |
| exit d15, no refill | 27 | 5.26 | 5.41 | +0.15 | 0.92 |

The per-arm offsets lie within ±0.2, so the two judge families agree on the effects as well as on the ordering. By domain, ρ is 0.82 (prose), 0.78 (code), 0.91 (hard); arithmetic (−0.03) and Chinese (0.49) have almost no rating variance to correlate.

## Appendix E. Draft-equals-target control on Llama-3.1-8B

To test whether the −0.14 reading of the lossless speculative arm on Llama arises from the speculative loop's mechanics (batched verification, cache cropping, carried-over tokens) rather than from the NF4 drafter, the loop was run with the drafter set to the unmodified bf16 model itself under the strict rule, on the 40 hard-verifiable and 40 prose prompts, and judged against the same reference pair. Acceptance was 7.8 of 8 drafted tokens per window (EAR 0.995), i.e. the floating-point floor rejects about 2.5% of tokens over these long responses.

| Hard-verifiable domain (n = 40) | Δ vs. reference pair | 95% CI |
|---|---|---|
| Speculative loop, drafter = unmodified model (control) | −0.16 | [−0.56, +0.24] |
| Speculative λ=1.0 (NF4 drafter) | −0.46 | [−0.81, −0.11] |
| Speculative λ=0.2 (NF4 drafter) | −0.44 | [−0.81, −0.07] |
| NF4 4-bit alone | −0.01 | [−0.46, +0.44] |
| Both NF4-drafted arms pooled (n = 80) | −0.45 | [−0.70, −0.20] |
| NF4-drafted arms minus control, paired | −0.29 | [−0.63, +0.06] |

On prose the control read +0.26 [+0.01, +0.51], within the range a 40-item cell produces (the second reference sample itself read +0.33 on prose). The loop mechanics therefore account for at most part of the hard-domain deficit; the remainder is not significant at this sample size and its cause was not determined.